# CHAPTER 7

# Machine Learning-based Biological Ageing Estimation Technologies: A Survey

Zhaonian Zhang[1], Richard Jiang[1], Danny Crookes[2] and Paul Chazot[3]

[1] School of Computing and Communication, Lancaster University, Lancaster, UK
[2] ECIT Institute, Queen's University Belfast, Belfast, UK
[3] School of Bioscience, Durham University, Durham, UK

## 1. Introduction

Aging of the population is an important challenge facing the world in the 21st century and has a profound impact on all aspects of society. As we age, the molecules, cells, tissues and organs in the human body change. It is very complicated to analyze the human aging process from a biological point of view [29], and no obvious features have been found so far. Generally speaking, aging is the gradual accumulation of harmful biological changes accompanied by the gradual loss of functions [21]. However, like most species, aging does increase the risk of morbidity and death in humans. Moreover, the process of human aging is different, especially the external manifestations of aging (for example, skin wrinkles, whitening of hair, cataract) are significantly different for each person and for age-related diseases, the age onset of each patient is also different. These show that age is not the only measure of aging, which has prompted scientists to work hard to measure aging from a biological point of view, with the goal of predicting human BA through 'aging biomarkers' and using BA as a standard for measuring aging. Compared with chronological age (CA), it can better predict remaining life and disease risk [4].

With the development of ML, BA's prediction technology has also made great progress. Computers can complete predictions quickly and accurately. Supervised ML covers many different algorithms that learn to recognize patterns and relationships between many input variables (features) in order to estimate one or more output variables (labels) as accurately and robustly as possible [17]. In this article, the input variables are different aging biomarkers, and the output variable is the predicted BA. The models are always training on healthy individuals, and in the analyzing process, we can compare the sample's predicted age and CA. If a sample's predicted age is older than his real age, it is thought to reflect poorer health with the risk of age-related diseases. If a sample's predicted age is younger than his real age, it shows that the subjects are healthy, have good living habits, pay attention to exercise and meditation, and may be highly educated [30, 40] (*see* Fig. 1).

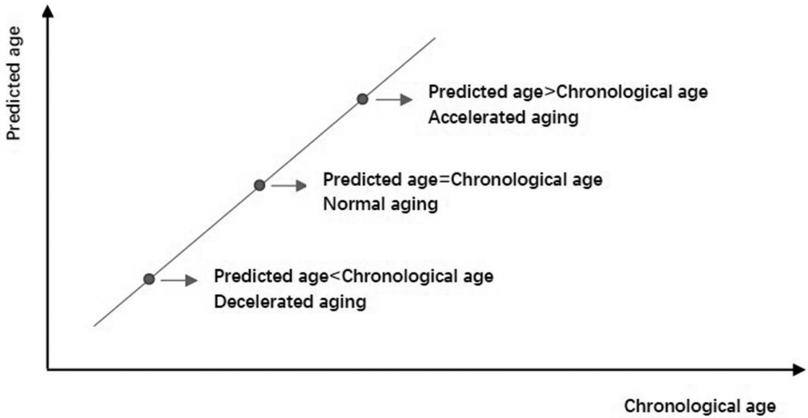

**Fig. 1:** Compare predicted age and chronological age

This technology can be used to predict the risk of age-related diseases, mortality and monitor biological aging over time. Not only for health detecting, but this technology can also play a very important role in research and development of medicine. At present, many pharmaceutical companies all over the world are committed to research of medicine for the treatment of age-related diseases, but generally speaking, the effect of these medications will not be obvious in the short term; even experienced doctors cannot judge whether the drugs have played a role in the short term, as it may take several years to follow up. This makes it very difficult for pharmaceutical companies to collect medicine data, which restrict the research and development of medicines for age-related diseases. However, this technology finds a new way to solve this problem. The models are always very sensitive; even if there are slight changes in the biomarkers, the models will show them up by the predicted age. This enables pharmaceutical companies to conduct follow-up investigations from the time the patients begin taking drugs, so as to know the effect of drugs in time and get patients' data quickly for further research.

Here, we will show the potential of supervized ML methods to estimate BA, focusing on the latest developments in computing a systemic (blood-based) brain-specific-age and facial-age estimation. Especially, due to the availability of blood test, brain imaging data and clinical events, these models show the best performance in terms of accuracy and prediction of mortality risk, and can be more easily applied to large longitudinal population studies. This article aims to provide references for researchers who are studying age-related diseases, and hopes to help them make contributions in this field.

## 2. ML for Blood Biomarkers

Blood is an important part of the human body. For a healthy person, the contents of various substances contained in the blood is within a certain range. If it is out of this range, it may mean a dangerous signal. So the blood test can detect a person's health and also the initial characteristics of certain diseases. In addition, as people age, the levels of specific markers in the blood usually fluctuate, such as glucose will increase and hemoglobin will decrease [39]. For these reasons, BA estimation based on the blood biomarker group has always been a research hotspot in aging research [39, 31, 25, 34, 36].

Recently, a method for estimating BA based on deep learning has been proposed. This innovative method is faster and more accurate than traditional methods. It uses healthy individuals as training data, blood circulation biomarkers as input features, and the sample's CA as a label [31, 25, 34, 36]. Deep learning is a specific branch of ML, which is usually used for identification, classification, and to explain the relationship between multiple variables in big data scenarios. Deep neural network is the most commonly used and most representative method in deep learning technology. It recognizes patterns in a large amount of data by imitating the structure and function of the brain. Generally, a deep neural network model has an input layer, one or more hidden 'decision-making' layers and output layers [6] (*see* Fig. 2).

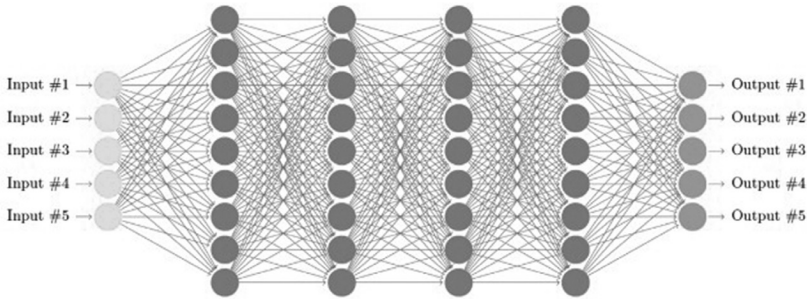

**Fig. 2:** The structure of deep neural network

In this article, the data of the input layer are the various blood biomarkers of the sample, and the output layer finally returns the CA predicted by the model. These algorithms can capture hidden underlying features and learn complex representations of highly multidimensional data [32]. Deep neural networks have powerful feature extraction capabilities, can capture hidden underlying features, and have strong fault tolerance. They perform better than ML on huge datasets.

### 2.1  Research Review on Blood Biomarkers

The first study about using blood biomarkers for BA prediction was conducted by Putin and colleagues [36. In this ground-breaking study, their data source was the anonymous blood biochemical records of 62,419 Russia's general population. Specifically, they used the ML method, with 41 standardized blood markers and sex as inputs and BA as outputs. Among all the algorithms, the deep neural network had the best performance in predicting BA. The final results showed that its average absolute error (MAE) was 6.07 years, and the Pearson correlation coefficient ($r$) between CA and BA was 0.9, coefficient of determination ($R^2$) was 0.8. The authors continued to experiment with the same data, by changing the hyperparameters of the deep neural network, such as the number of hidden layers, the number of neurons in the hidden layer, the 21 best performing models were obtained. And finally, they built an ensemble model through these models, whose performance has been greatly improved, MAE=5.55, $r$=0.91, and $R^2$ =0.83. After that, they tested the model with a core set of the 10 most predictive circulating biomarkers. The final result showed that the accuracy of the model was still very good ($R^2 = 0.63$), which indicated that the model has strong robustness [36].

Mamoshina *et al*. [31] used similar algorithms for experiments, but they selected three datasets for the population of specific countries to train models. Specifically, they are the Canadian dataset containing 20,699 samples, the South Korean dataset containing 65,760 samples, and the East European dataset containing 55,920 samples. The researchers used 19 blood biomarkers and gender as input data for the model, and age as the label. The model was trained in each dataset and tested on independent test sets of all available populations. The results showed that when these models were trained and tested on the same population, they showed great performance with their MAE ranging from 5.59 to 6.36 years, and $R^2$ ranging from 0.49 to 0.69. However, when these models were trained in one population and tested in another population, the performance of the models decreased significantly, with their MAE ranging from 7.1 to 9.77 years, and $R^2$ ranging from 0.24 to 0.34. Similarly, when the neural network used all the data combined from the three datasets for training, including the population label as additional feature, the performance of the model was greatly improved. Researchers used the single population to test this model (each of the three datasets was tested), found its MAE ranging from 5.60 to 6.22 years, and its $R^2$ ranging from 0.49 to 0.70. When the model was tested on the combined dataset, MAE = 5.94 and $R^2$ = 0.65 (*see* Table 1).

Table 1: Biological age estimates based on blood biomarkers

| Input features | Algorithm | Population | Num (training:test) | MAE | Pearson $r$ | $R^2$ |
| --- | --- | --- | --- | --- | --- | --- |
| 41 blood biomarkers | DNN | Russian | 62,419 (90:10) | 6.07 | 0.9 | 0.8 |
| 19 blood biomarkers | DNN | Canadian | 20,699 (80:20) | 6.36 | 0.7 | 0.52 |
| | | South Korean | 65,760 (80:20) | 5.59 | 0.7 | 0.49 |
| | | Eastern European | 55,920 (80:20) | 6.25 | 0.84 | 0.69 |
| | | All | 142,379 (80:20) | 5.94 | 0.8 | 0.65 |

The above results indicated that blood biomarkers have population specificity [31], which may be due to the different environments of populations, or genetic differences caused by different ancestors. Through feature significance analysis, the five features that had the greatest impact on the results were gender, albumin, glucose, hemoglobin and urea levels. In addition, the content of blood biomarkers was related to a person's age and whether they have age-related diseases [7], and the prediction results for female age group were more accurate than that for men [31].

It is worth noting that a recent study by Mamoshina *et al*. [33] proved a positive correlation between smoking and BA. For women, the BA estimate for smokers was twice that of non-smokers, while for men it was 1.5 times. The difference was significant under age 40 and continued until age 55; subsequently, it gradually disappeared, possibly because of the increasing ability of smokers to resist the dangers of smoking after age 55 [33].

## 3. ML for Understanding Brain Age

Brain Age (BrA) is also a very important kind of BA. The aging brain functions decline and neurodegenerative diseases bring increasingly serious economic, old-age, medical, and other social problems to our society. Therefore, it is an important task for researchers to accurately and quickly predict the BrA of subjects. Although brain aging is a natural process, there are significant individual differences in changes in brain volume, cortical thickness and white matter microstructure during this process. In addition, the deviation degree between the individual brain aging trajectory and the average trajectory of healthy brain aging can reflect the individual's future risk of neurodegenerative diseases [7, 15]. Therefore, building models based on the characteristic patterns of brain aging contained in neuroimaging data and detecting the aging trajectories of individual brains can provide a new perspective for studying individual differences in brain aging.

In previous studies, researchers have built various models with different types of data. The common types of data include sMRI (structural magnetic resonance imaging), fMRI (functional magnetic resonance imaging), and DTI (diffusion tensor imaging) (*see* Fig. 3). Here, we will focus on the most recent and accurate developments about the research based on sMRI [7, 8, 9, 10, 11, 12, 13].

There are two main algorithms – one is Gaussian Process Regressions and the other is Convolutional Neural Networks (CNN). GPR obtains the NxN similarity matrix by normalizing gray matter (GM) and white matter (WM) images, and then predicts BA through regression tasks, and makes a large number of features conform to the multivariate Gaussian distribution. This algorithm can reflect the local pattern of covariance between individual data points. As for CNN, it is a class of deep neural network that has proved very powerful in image classification (*see* Fig. 4). CNN has many advantages. It can effectively learn the corresponding features from a large number of samples, avoiding the complicated feature extraction process. It uses a simple non-linear model to extract more abstract features from the original image, and only a small amount of human involvement is required in the whole process. CNN has two characteristics – local perception and parameter sharing. Local perception makes each neuron of the CNN to avoid the need to perceive all the pixels in the image, but only perceive the local pixels of the image, and then combine these in a higher layer. The partial information is merged to obtain all the characterization information of the image. The neural units of different layers are locally connected, that is, the neural units of each layer is only connected to some of the neural units of the previous layer. Each neural unit only responds to the area in the receptive field and does not care about the area outside the receptive field at all. Such a local connection mode ensures that the learned convolution kernel has the strongest response to the input spatial local mode. The weight-sharing network structure makes it more similar to a biological neural network, which reduces the complexity of the network model and reduces the number of weights. This kind of network structure is highly invariant to translation, scaling, tilt, or other forms of deformation. Moreover, CNN uses the original image as input, which can effectively learn the corresponding features from a large number of samples, avoiding the complicated feature extraction process. Here, the input data is brain sMRI, and output is the predicted BA.

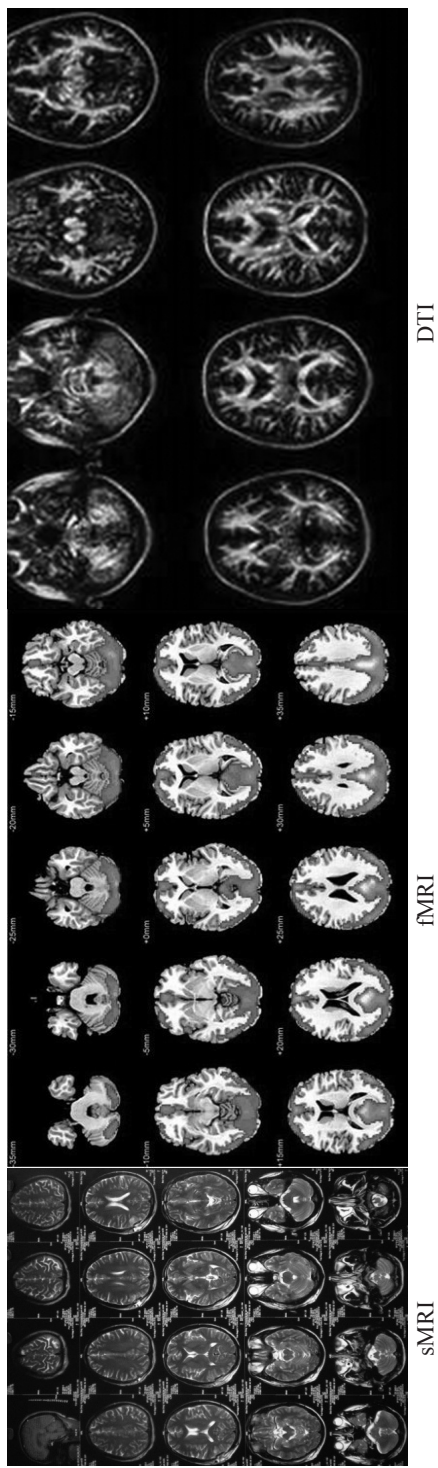

**Fig. 3:** Examples of sMRI, fMRI and DTI

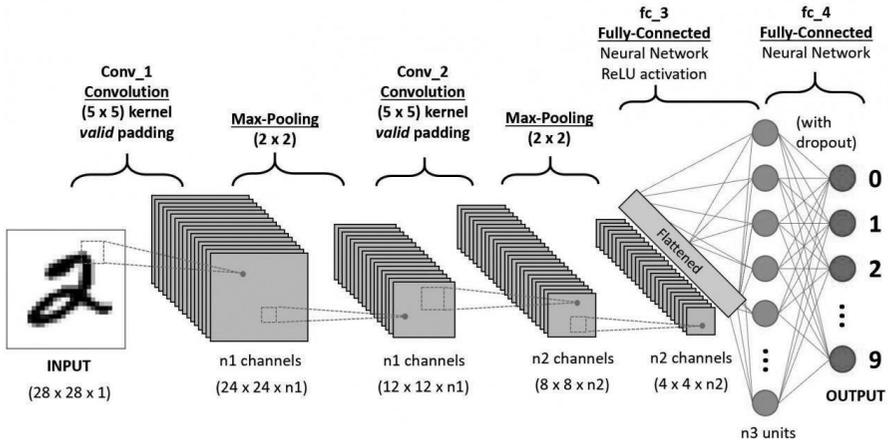

**Fig. 4:** Structure of a CNN

## 4. Research Review on BrA

A recent study showed that BrA was relatively accurate in predicting BA after experiments performed by using a large healthy control dataset with a sample size of 2001. When using the GPR algorithm, if the data normalized GM images, the test results showed MAE = 4.66 and $R^2$ = 0.89. The results of using the CNN algorithm were slightly better, with MAE of 4.16 and $R^2$ of 0.92. When the data uses normalized WM images, the results of the two algorithms were also similar. The GPR algorithm had an MAE of 5.88 and $R^2$ of 0.84, compared with CNN's MAE of 5.14 and $R^2$ of 0.88. But when raw (non-parcelated) data was used, the gap between the two algorithms was large. GPR had an MAE of 11.81 and $R^2$ of 0.32, while CNN still maintained good performance with MAE of 4.65 and $R^2$ of 0.88. Interestingly, when using the combined data of GM and WM, both algorithms achieved the best performance. Whether it was GPR or CNN, the results said that the value of Intraclass

Correlations Coefficients was very high (>0.9), which showed high levels of within-scanner and between-scanner reliability in BrA estimation (*see* Table 2).

**Table 2:** Biological Age Estimates Based on Brain sMRI

| Input features | Algorithm | Num (training:test) | MAE | Pearson r | $R^2$ |
|---|---|---|---|---|---|
| Structural MRI (normalized GM images) | GPR | 2,001 (90:10) | 4.66 | 0.95 | 0.89 |
| Structural MRI (normalized WM images) | | | 5.88 | 0.92 | 0.84 |
| Structural MRI (raw data) | | | 11.81 | 0.57 | 0.32 |
| Structural MRI (normalized GM+WM images) | | | 4.41 | 0.96 | 0.91 |
| Structural MRI (normalized GM images) | CNN | | 4.16 | 0.96 | 0.92 |
| Structural MRI (normalized WM images) | | | 5.14 | 0.94 | 0.88 |
| Structural MRI (raw data) | | | 4.65 | 0.94 | 0.88 |
| Structural MRI (normalized GM+WM images) | | | 4.34 | 0.96 | 0.91 |

According to a study by Imperial College London, the greater the difference between predicted BrA and CA in older people, the higher the risk of mental or physical problems and early death. Researchers found that the brains of patients suffering from Alzheimer's disease, traumatic brain injury, and psychosis usually accelerate aging [10, 18, 20]. In 2016, Steffener, Luders and others proved that education, meditation, and physical exercise can make the brain young and energetic [30, 40]. In addition, Franke K. and her colleagues [19] studied the relationship between diet and BrA. By comparing healthy groups, they observed that the brains of baboons who experienced malnutrition tended to age prematurely. More recently, Hatton and colleagues [22] reported an association between negative fateful life events and advanced brain aging after controlling for physical, mental, and lifestyle factors.

## 5. ML for Facial Image Processing

Facial features are the parts of the face that contain the most information. Facial features generally include eyes, nose, mouth, facial wrinkles, as well as more complex attributes, such as gender and emotion. Among them, gender, race, age and other characteristics can be extracted by low-resolution photos [23], but details, such as moles and facial scars need high spatial frequency images to be identified [24].

Facial age is also a kind of BA, the estimation of which is one of the most important projects in the future of computer science applications. However, estimation systems face many challenges. For example, first, the data for each age is not evenly distributed [3, 26]. Secondly, the difference between images is huge, the resolution of the pictures, the light, the posture of people, etc. are all different [26, 38, 27, 35]. Thirdly, aging of the human face is affected by many external factors, such as living environment, diseases and other factors [1]. Finally, the face is also directly related to factors, such as gender and race [16]. Therefore, after continuous attempts by researchers, several methods to improve the accuracy of facial age prediction models have been summarized. The first thing is that it is better to train a model with a constrained dataset than an unconstrained dataset. Second, in addition to facial images, it will be better to add gender and race as the input data. The third is that in the choice of CNN, the more the layers, the higher the accuracy of the model. In addition, the age range of the data should be wide, so that the sample distribution of the training set is more balanced. Finally, before training the model, the pictures can be preprocessed, such as by aligning faces.

The age estimation system sequence is shown in Fig. 5. In general, the two most important parts of the whole process are feature extraction and age estimation. The face detection step before this is to ensure that the input data are face objects. The subsequent face alignment step can improve the prediction accuracy of the model. Fig. 6 shows the meaning of face detection and face alignment.

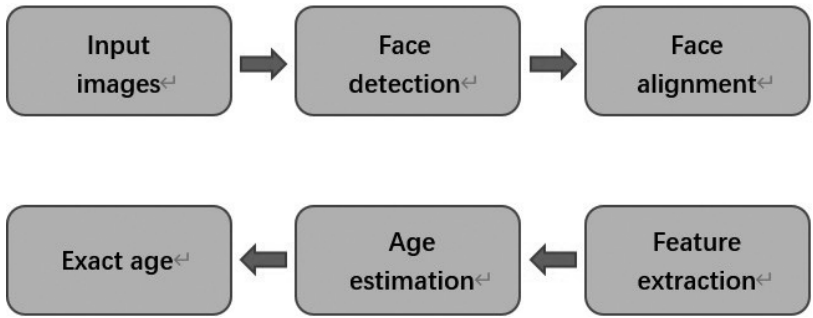

**Fig. 5:** The process of facial age estimation

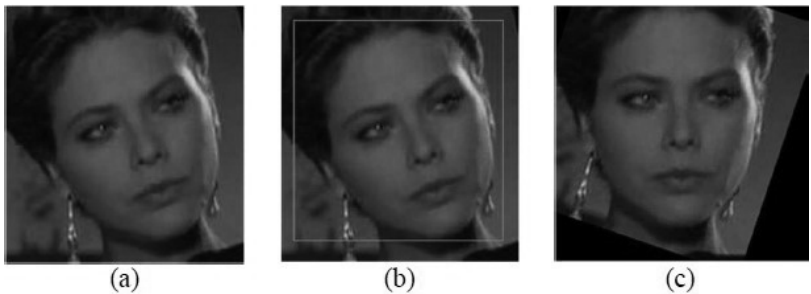

**Fig. 6:** (a) input image, (b) face detection, (c) face alignment

After this, the computer will extract the features of the input picture, and finally give the predicted age classification or calculation result. Feature extraction generally has two methods, namely, manual algorithm and deep learning. Manual algorithm can predict age even if the training data is small, but its accuracy is not high, and it is necessary to manually define the features to be extracted in advance. Deep learning needs to be trained on a large amount of data, but it has high accuracy and can automatically extract features.

As for age estimation, there are four basic methods – classification, regression, ranking, and hybrid. Classification is suitable for data with few age categories, but not for data with accurate and continuous ages. And for each age group, it needs a large number of samples. The regression is more suitable for cases with target exact age. However, if the age distribution of samples is unbalanced, it will be more sensitive and less robust than other methods. Ranking is very suitable for unbalanced data distribution, but its accuracy is the worst among all the methods. Hybrid is the combination of classification and regression, so its performance is the best, but correspondingly, it has higher complexity.

## 6. Research Review on Facial Age Estimation

The first study on facial age estimation was completed by Kwon and Lobo in 1994. They used the algorithm of Snakelets + classification to classify facial images into age groups (baby, young, old) on a small dataset which they collected by themselves. The accuracy of the final model was 100% [14]. In 2015, Ali *et al.* used Canny Edge Detection + Classification to divide the age of the sample into three groups (10-30, 30-50 and 50+) by three regions of the face image (cheeks, eye corners and forehead). The accuracy of the final model ranged from 62.86 to 72.48 [2]. Wan *et al.* used two datasets for the experiment, and added sex and race as additional conditions to the CNN model to estimate the exact age. The final MAE was 2.93 for the constrained dataset and 5.22 for the unconstrained dataset [42]. In 2019, Xie *et al.* used two pre-trained CNN models (Alexnet, VGG-16) to create a framework that uses age-related facial attributes to estimate the age of humans. The results said that on the constrained dataset, MAE was 2.69; on the unconstrained dataset, MAE was 5.74 [43]. In a recent study by Liu *et al.*, two datasets were used to test the accuracy of the multi-task learning + hybrid to predict the true age. The results showed that MAE ranged from 2.30 to 3.02 on the constrained dataset, and 5.67 on the unconstrained dataset [28] (*see* Table 3).

## 7. Conclusion

This article reviews the development of projects using ML for age prediction. From the perspective of medical and clinical values, experiments based on blood biomarkers and brain sMRI are more valuable, but facial-based age prediction can also provide help for advanced technologies, such as face recognition. Now, with the exploration of deep learning, more and more algorithms have been developed, which provide powerful tools for research in various fields. But these algorithms

**Table 3:** Facial Age Estimation Research

| Research | Algorithm | Dataset | Target | Accuracy (%) | MAE |
|---|---|---|---|---|---|
| Kwon and Lobo, 1994 | Snakelets + classification | 47 images | Age group | 100 | – |
| Ali *et al.*, 2015 | Canny edge detection+ classification | 885 images | Age group | 62.86-72.48 | – |
| Wan *et al.*, 2018 | CNNs+regression | Morph II CACD | Real age | – | 2.93 5.22 |
| Xie *et al.*, 2019 | Alexnet, VGG-16 +classification | Morph II AgeDB | Real age | – | 2.69 5.74 |
| Liu *et al.*, 2020 | Multi-task learning+hybrid | Morph II | Real age | – | 2.30– 3.02 |
|  |  | UTKFace |  |  | 5.67 |

*Note:*
**Morph II** is a constrained dataset containing 55134 images, the age range is from 16 to 77, 46,645 males and 8,489 females [37].
**CASD** is an unconstrained dataset containing 163446 images, the age range is from 16 to 62. It also has additional information such as the name of the celebrity, date of birth, and estimated year of which the photo was taken [5].
**AgeDB** is an unconstrained dataset containing 16488 images, the age range is from 1 to 101, 9788 males and 6700 females [43].
**UTKFace**'s number of total images>20000, it is an unconstrained dataset containing, the age range is from 0 to 116. It also has additional information such as gender, ethnicity, and datetime of each image [41].

actually still have limitations. First, the process of ML giving results is invisible, and it is difficult to explain with mathematical or biological knowledge. Secondly, adjusting the parameters of the model is a complex process, which generally needs experienced people to do. Finally, the input data is likely to have errors, possibly due to the instruments or other factors.

In future, as it becomes easier to obtain data, researchers can shift their focus to BA prediction for specific organs. For example, predicting the age of the lungs by vital capacity, or predicting the age of the heart by the frequency of heartbeats, this kind of research has important value for analyzing diseases of specific organs.

Overall, we want to help people who are currently studying BA predictions through a summary of existing technologies, and hope that in future, we can take part in the big data revolution together.